%% file: main.tex
\documentclass[conference]{IEEEtran}
\IEEEoverridecommandlockouts
\usepackage{cite}
\usepackage{amsmath,amssymb,amsfonts}
\usepackage{algorithmic}
\usepackage{graphicx}
\usepackage{textcomp}
\usepackage{xcolor}

\usepackage{xac}

\def\BibTeX{{\rm B\kern-.05em{\sc i\kern-.025em b}\kern-.08em
    T\kern-.1667em\lower.7ex\hbox{E}\kern-.125emX}}
\begin{document}

\title{Supporting Mitosis Detection AI Training with Inter-Observer Eye-Gaze Consistencies}

\author{

\IEEEauthorblockN{Hongyan Gu}
\IEEEauthorblockA{\textit{Electrical and Computer Engineering} \\
\textit{University of California, Los Angeles}\\
Los Angeles, USA \\
ghy@ucla.edu}
\and
\IEEEauthorblockN{Zihan Yan}
\IEEEauthorblockA{\textit{Informatics Programs} \\
\textit{University of Illinois, Urbana-Champaign}\\
Urbana, USA \\
zihan25@illinois.edu}
\and
\IEEEauthorblockN{Ayesha Alvi}
\IEEEauthorblockA{\textit{Electrical and Computer Engineering} \\
\textit{University of California, Los Angeles}\\
Los Angeles, USA \\
aalvi@ucla.edu }
\and
\IEEEauthorblockN{Brandon Day}
\IEEEauthorblockA{\textit{Electrical and Computer Engineering} \\
\textit{University of California, Los Angeles}\\
Los Angeles, USA \\
bjday@ucla.edu }
\and
\IEEEauthorblockN{Chunxu Yang}
\IEEEauthorblockA{\textit{Electrical and Computer Engineering} \\
\textit{University of California, Los Angeles}\\
Los Angeles, USA \\
chunxuyang@ucla.edu}
\and
\IEEEauthorblockN{Zida Wu}
\IEEEauthorblockA{\textit{Electrical and Computer Engineering} \\
\textit{University of California, Los Angeles}\\
Los Angeles, USA \\
zdwu@ucla.edu}
\and
\IEEEauthorblockN{Shino Magaki}
\IEEEauthorblockA{\textit{Pathology and Laboratory Medicine} \\
\textit{UCLA David Geffen School of Medicine}\\
Los Angeles, USA \\
smagaki@mednet.ucla.edu}
\and
\IEEEauthorblockN{Mohammad Haeri}
\IEEEauthorblockA{{\textit{Pathology and Laboratory Medicine}} \\
\textit{Kansas University Medical Center}\\
Kansas City, USA \\
mhaeri@kumc.edu}
\and
\IEEEauthorblockN{Xiang `Anthony' Chen}
\IEEEauthorblockA{\textit{Electrical and Computer Engineering} \\
\textit{University of California, Los Angeles}\\
Los Angeles, USA \\
xac@ucla.edu}

}

\maketitle

\input{00_abstract}


\input{01_introduction}
\input{02_method}
\input{03_result}
\input{04_discussion}

\section*{Acknowledgment}

This work was funded by the Young Investigator Award by the Office of Naval Research. We would like to thank Dr. Harry V. Vinters, Christopher Kazu Williams, Dr. Sallam Alrosan, Dr. Issa Al-kharouf, and Ellie Onstott for their assistance in developing the mitosis dataset. We appreciate the anonymous reviewers for their comments in improving the manuscript, as well as participants for their time and participation.

\bibliographystyle{IEEEtran}
\bibliography{references}

\end{document}

%% file: 00_abstract.tex
\begin{abstract}

The expansion of artificial intelligence (AI) in pathology tasks has intensified the demand for doctors' annotations in AI development. However, collecting high-quality annotations from doctors is costly and time-consuming, creating a bottleneck in AI progress. This study investigates eye-tracking as a cost-effective technology to collect doctors' behavioral data for AI training with a focus on the pathology task of mitosis detection. One major challenge in using eye-gaze data is the low signal-to-noise ratio, which hinders the extraction of meaningful information. We tackled this by levering the properties of inter-observer eye-gaze consistencies and creating eye-gaze labels from consistent eye-fixations shared by a group of observers. Our study involved 14 non-medical participants, from whom we collected eye-gaze data and generated eye-gaze labels based on varying group sizes. We assessed the efficacy of such eye-gaze labels by training Convolutional Neural Networks (CNNs) and comparing their performance to those trained with ground truth annotations and a heuristic-based baseline. Results indicated that CNNs trained with our eye-gaze labels closely followed the performance of ground-truth-based CNNs, and significantly outperformed the baseline. Although primarily focused on mitosis, we envision that insights from this study can be generalized to other medical imaging tasks.
\end{abstract}

%% file: 01_introduction.tex
\section{Introduction}

Recently, artificial intelligence (AI) has shown exciting advancements in analyzing pathology specimens and assisting diagnoses \cite{cui_artificial_2021, niazi2019digital}. However, developing pathology AI may face bottlenecks in acquiring high-quality annotations, a process that is both time-consuming and challenging due to the limited availability of pathologist annotators \cite{montezuma_annotating_2023}. In light of this, there has been a growing interest in collecting eye-gaze data from doctors to train medical AI models \cite{lim2022eye}. The advantage of the eye-gaze approach is its cost-efficiency \cite{9706338}: the eye-gaze collection process can be integrated into the doctors' workflow without requiring them to spare additional time for annotations. To date, eye-gaze data have shown promising potential in training AI models for radiology \cite{karargyris2021creation, 10030981, stember2020integrating}, retinopathy \cite{jiang2023eye, jiang2024dcamil}, and pathology tasks \cite{9706338}.

Eye-gaze data can be used as additional information to complement ground-truth labels. This approach has been particularly effective in chest X-ray image classification, where combining class labels with radiologists' eye-gaze data has resulted in improved model performance\cite{karargyris2021creation, wang2022follow, saab2021observational, 10.1145/3554944.3554952, 10030981, 10155473}. Additionally, eye-gaze data can also be directly employed as training labels. For example, Stember \etal collected masks of meningiomas in radiology images by instructing radiologists to explicitly move their eyes around the contours and subsequently trained a segmentation model \cite{stember2019eye}. However, this approach did not fully align with the routine of radiologists. In a subsequent study, Stember \etal \cite{stember2020integrating} simulated the environment of real clinical interpretations and collected both speech and eye-gaze data from radiologists. They further trained AI with both modalities and achieved an 85\% test accuracy in predicting lesion locations.

Despite the improvements in other medical domains, the application of eye-gaze data in developing pathology AI is still sparse and encounters significant challenges. Firstly, pathology tasks often involve detecting objects, which requires additional localization information compared to classification tasks. Secondly, pathology image data tend to be less structured and exhibit higher variation. Finally, the inherently low signal-to-noise ratio of eye-gaze data poses a challenge in extracting meaningful information. A pioneering work of pathology AI + eye-gaze labels was conducted by Mariam \etal \cite{9706338}. Their research focused on detecting keratin pearls in squamous cell carcinoma pathology images. Notably, they reported that the performance of the eye-gaze-trained detection model was comparable to those trained with ground truth labels. However, their approach required pathologists to fixate on each region of interest for one to two seconds to mitigate noise in the eye-gaze data. Similar to \cite{stember2019eye},  this eye-gaze collection method did not match with pathologists' natural viewing behavior.

This work introduces and evaluates a pipeline that enables Convolutional Neural Networks (CNNs) to learn from humans' eye-gaze behaviors without disrupting the routine of pathology image reading. Different from \cite{9706338}, we distilled meaningful information from the noisy eye-gaze data according to the inter-observer eye-gaze consistencies -- by extracting consistent eye-fixations shared among a group of human participants (\ie consistent fixations). This approach was inspired by \cite{brunye2021melanoma}, which demonstrated that these consistent fixations could align with locations of critical histological patterns. We focused on a pathology task of mitosis detection in the brain tumor of meningiomas, which is critical for meningioma grading \cite{10.1093/neuonc/noab106, 10.1145/3577011}.

For the experiment, we collected eye-gaze data from 14 non-medical participants viewing 800 meningioma images. These non-medical participants underwent a training and screening session to ensure the quality of collected eye-gaze data. We generated eye-gaze labels with data from 3 -- 14 participants and assessed the label quality. We found including more participants could improve the quality of eye-gaze labels, with observed improvements in precision and reduction in the variation of recall. Furthermore, we trained CNNs based on labels generated by three approaches: \one a heuristic-based approach that relied on color information \cite{ruifrok2001quantification}, \two eye-gaze data, and \three ground truth. These CNNs were tested on a Whole Slide Image (WSI) of meningioma, including 1,298 non-background High-Power Fields (HPFs) (approximately $3.3\times 10^9$ pixels) and 380 mitoses. The heuristic-based, eye-gaze-based, and ground-truth-based CNNs achieved average precision scores of 0.79, 0.83, and 0.85, respectively. The average recall for CNNs of the three conditions were 0.47, 0.61, and 0.72. Visualizations of the precision-recall curves indicated that the performance of eye-gaze-based CNNs closely followed those trained with ground truth, and significantly outperformed the heuristic-based ones.

%% file: 02_method.tex
\section{Materials and Methods}

This study had been approved by the Institutional Review Board of the University of California, Los Angeles (IRB\#21-000139) and was conducted between September 2023 and December 2023. It consists of three parts: \one collection of specimens and pathology images; \two conducting a user study and acquiring participants' eye-gaze data; \three training CNNs based on eye-gaze data and its evaluation.

\begin{figure*}
    \centering
    \includegraphics[width=0.995\linewidth]{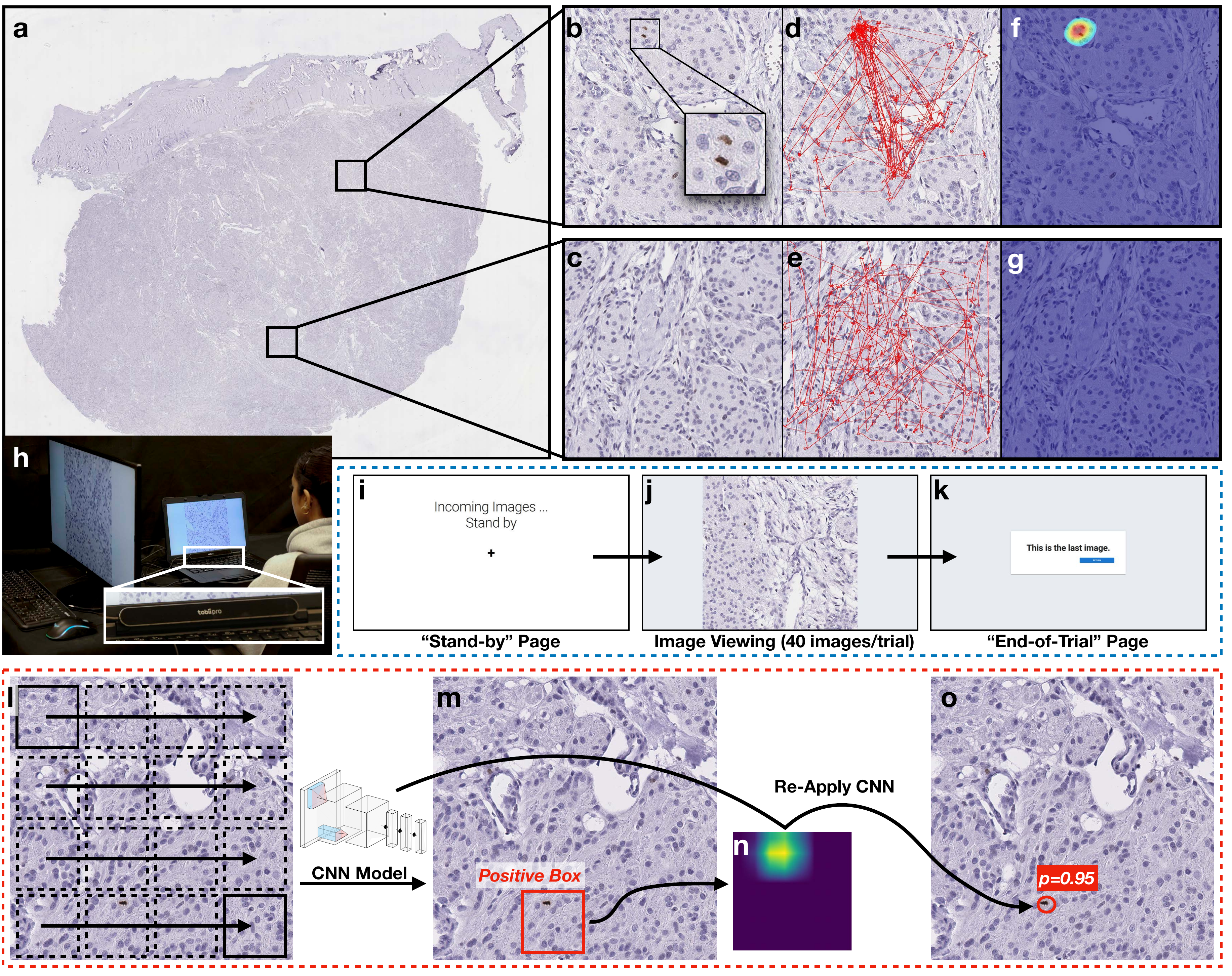}
    \caption{Methods: (a) A meningioma Whole Slide Image used for sampling the 1,000 HPF images. The specimen was stained with PHH3 immunohistochemistry.  A (b) positive and (c) negative HPF image in the 1,000 HPF image collection. (d, e) Visualizations of 14 participants' eye-gaze sequences while viewing the HPF images of (b) and (c). (f, g) The corresponding eye-gaze heatmaps after processing. (h) Apparatus for eye-gaze user study. The participant was seated on the right side and the moderator who controlled the experiment was on the left. Screenshots of the image viewing interface used in the eye-tracking sessions: (i) the ``Stand-by'' page shown at the beginning of each trial; (j) in each trial, the system would display 40 HPF images in random order and with random image transforms; (k) the ``End-of-Trial'' page marking the end of each trial. Pipeline enabling CNNs to predict mitosis locations with probabilities: (l) the sliding window CNN was applied; (m) boxes with positive CNN classifications were retained to generate the (n) saliency map. The centroid of saliency map hotspots were used as the locations of detected mitoses. (o) The same CNN was re-applied to each location to calculate the probability.}
    \label{fig:method}
\end{figure*}

\subsection{Pathology Image Acquisition}
\label{s3p1}

We collected 1,000 meningioma images stained with Phospho-histone H3 (PHH3) immunohistochemistry \cite{10.1093/neuonc/nov002}, including 500 positive images (with mitoses, Figure \ref{fig:method}(b)) and 500 negative images (without mitoses, Figure \ref{fig:method}(c)). These images were randomly sampled from two meningothelial-transitional meningioma whole slide images (WSIs, Figure \ref{fig:method}(a)) of one patient\footnote{The WSIs were scanned with Aperio CS2 scanner, 400$\times$ magnification, 0.25$\mu m$ per pixel. Non-tumor regions were excluded from the sampling.}. Each image had a size of 1,600$\times$1,600 pixels (or 400$\times$400$\mu m$), which was equivalent to an HPF under 400$\times$ magnification in the light microscopy. Furthermore, we selected one test WSI containing 1,298 non-background HPFs for evaluation. The test WSI was acquired from another patient. This WSI selection process seeks to create biological independence between training/validation and test sets, which can create a more generalized setting to test the adaptivity of AI.

Out of the 1,000 HPF images, 120 were used for participant training, 40 for participant screening (Section \ref{s3p2}), and 800 for collecting participants' eye-gaze data (Section \ref{s3p3}).

Three pathology professionals annotated the mitosis ground truth: two pathology residents (post-graduate year 3) independently reviewed all images first, followed by a third board-certified neuropathologist reviewing these initial annotations and providing the final ground truth \cite{gu_enhancing_2024}. The 1,000 selected HPFs had 705 mitoses. The test WSI included 380 mitoses.

\subsection{Participant Screening}
\label{s3p2}

Due to the challenges in recruiting medical professionals, we opted to recruit participants without requiring a medical background. To ensure their familiarity with the mitosis detection task, each participant was mandated to join a training session first (average length: 31 minutes). Participants' skills of mitosis detection were evaluated by a post-training survey, where they reported mitoses of 40 HPF images (matching the format of the eye-tracking sessions). To ensure the quality of eye-gaze data, only those who achieved an F1 score above 0.75 were identified as qualified participants and were invited to participate in the eye-tracking sessions.

We recruited 20 participants from a local university, including 17 undergraduate and three graduate students (ages ranging between 18 and 27). All participants had normal or corrected-to-normal vision. Of these, 6/20 participants did not pass the post-training screening, leaving 14/20 qualified participants for eye-tracking sessions.

\subsection{Apparatus for Eye-Tracking Sessions}
\label{s3p3}

Participants' eye movements were captured by a Tobii Pro Nano remote eye-tracker at 60Hz. Participants were seated approximately $60.4cm$ away from the eye-tracker. The eye-tracker was attached to a 14-inch Dell Latitude 5430 laptop with 1920$\times$1080 monitor resolution (Figure \ref{fig:method}(h)).

A web-based interface was implemented to display the 800 HPF images to participants. This interface could show images in a random order. For each image, the interface could perform random image transforms, including rotations (0$^\circ$, 90$^\circ$, 180$^\circ$, or 270$^\circ$) and flips (horizontal or vertical), to reduce spatial estimation errors from the eye-tracker \cite{10.1145/3025453.3025599}.

\subsection{Procedure for Eye-Tracking Sessions}
\label{s3p4}

We split the 800 HPF images in the eye-tracking sessions into 20 trials (\ie trial \#1 to trial \#20), each including 20 positive and 20 negative images. We set the length of a trial as 40 images to avoid causing participants' fatigue or impacting eye-tracking performance. A moderator guided each participant through each trial, following the steps below:

\begin{itemize}
    \item \textbf{Eye-Tracker Calibration}: Each trial began with a nine-point calibration of the eye-tracker.
    \item \textbf{Stand-by Page}: A brief pause, during which participants were directed to focus on the center of the screen, prepping them for the subsequent images (Figure \ref{fig:method}(i)).
    \item \textbf{Image Viewing}: For each of the 40 images in a trial:
        \begin{itemize}[wide=\dimexpr\parindent+\labelsep\relax, leftmargin=* ]
        \item The system randomly selected an image (without replacement) and applied random image transforms (Section \ref{s3p3}) (Figure \ref{fig:method}(j)).
        \item Participants reported the number of mitoses in each image to the moderator.
        \item The moderator controlled the interface to display the following image.
    \end{itemize}
    \item \textbf{End-of-Trial Page}: Would be shown if all images had been viewed (Figure \ref{fig:method}(k)).
\end{itemize}

To reduce the ordering effect, participants assigned with an odd ID number underwent the trials in forward order, while those with an even ID number proceeded in reverse. Typically, participants completed all 20 trials over two one-hour visits, with ten trials each.

\subsection{Training CNNs from Eye-Gaze Data}
\label{s3p5}

For each image, we used the following steps to acquire consistent eye-gaze fixations shared by a group of $k$ participants.

\begin{enumerate}
    \item Gaze-point series from a group of $k$ participants were projected to the image according to the image transform record (\ie how the image had been rotated or flipped). Gaze points off the image, of low confidence, or `N/A' values were discarded (Figure \ref{fig:method}(d), (e)).
    \item A Gaussian kernel with a radius of 30 was applied to all projected gaze-points to create a gaze heatmap. High-heat areas could represent consistent fixations across the $k$ participants. To reduce noise, values below a threshold of $0.0018k$ in the gaze heatmap were set to 0. Additionally, hotspots with sizes smaller than 400 pixels were removed (Figure \ref{fig:method}(f), (g)).
    \item A blob detection algorithm identified the centroid of each remaining hotspot, which served as eye-gaze labels.
\end{enumerate}

We trained EfficientNet-b3 CNN models \cite{pmlr-v97-tan19a} using these eye-gaze labels. A two-iteration active learning approach was employed: In the first iteration, the model was trained on positive patches and randomly sampled negative patches\footnote{All patches mentioned in this work had a size of 240$\times$240 pixels.}. The second iteration served as a hard-negative mining process, which involved retraining with marginal patches identified by the first-iteration model. Because CNNs could not predict locations directly, we used a pipeline in \cite{10.1007/978-3-031-33658-4_21} to acquire the locations of detected mitoses with their probabilities:

\begin{enumerate}
\item \textbf{Sliding Window CNN:} The CNN was slid (step size = 60) on pathology images to classify positive boxes.
\item \textbf{Saliency Map Generation:} For each positive box, a saliency map was generated by the GradCAM++ \cite{gradcam++}.
\item \textbf{Location Extraction:} The centroid of the hotspots in the saliency map was used as the prediction location.
\item \textbf{Probability Assessment:} The same CNN was reapplied to each predicted location to calculate mitosis probability.
\end{enumerate}

The training set consists of 80\% of the 800 HPF images in the eye-tracking sessions, while the remaining 20\% were utilized for validation and threshold tuning. In the first iteration, the model was initialized with ImageNet weights and was trained using an SGD optimizer (momentum = 0.9, weight decay = $1\times 10^{-4}$), cross-entropy loss, and a Cosine Annealing learning rate scheduler with restart (max learning rate = 0.001, cycle = 9 epochs) for 18 epochs. The second iteration training started from the checkpoint with the highest validation F1 score in the first iteration and was trained with the same hyperparameters.

\subsection{Measurements \& Conditions}

For the evaluation, we measured both the quality of eye-gaze labels and their efficacy for model training.

\subsubsection{Quality of Eye-Gaze Labels} For the 800 HPF images viewed in the eye-tracking sessions, we assessed the quality of eye-gaze labels as a function of group size $k$, ranging from 3 to 14. For each $k$, we ran the eye-gaze label generation 20 times, with each run randomly selecting $k$ participants. The eye-gaze labels were compared against the ground truth, and the quality was measured by the precision, recall, and F1 score.

\begin{figure*}
    \centering
    \includegraphics[width=1.0\linewidth]{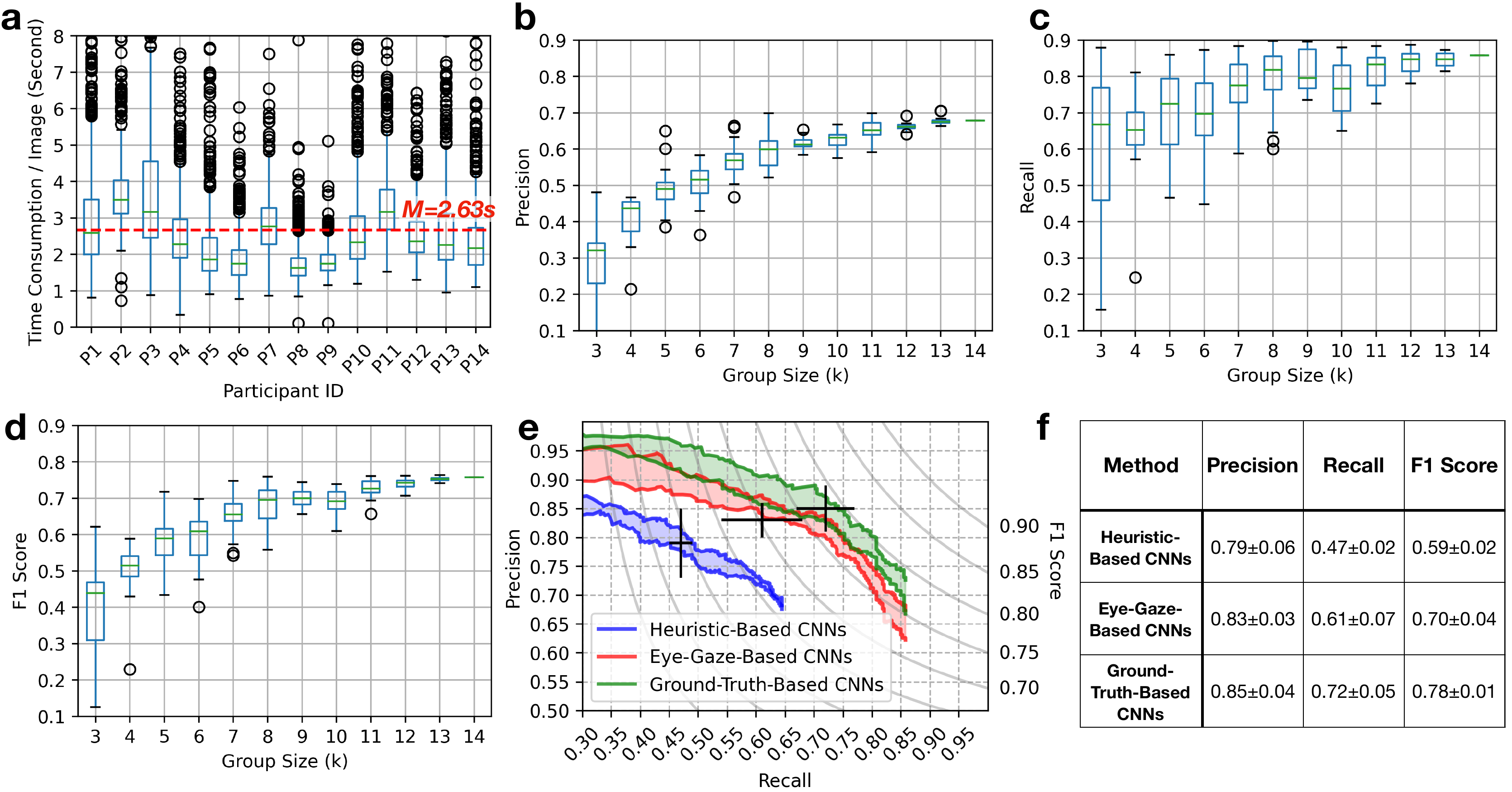}
    \caption{Experiment results: (a) Box-whisker plot of time consumption of the 14 qualified participants (\ie P1 -- P14) viewing images in the eye-tracking sessions. The (b) precision, (c) recall, and (d) F1 scores of the eye-gaze labels for the 800 HPF images in the eye-tracking sessions. (e) Ranges of precision-recall curves of the EfficientNet-b3 CNNs on the test WSI. The CNNs were trained from heuristic-based labels, eye-gaze labels ($k$=14), and ground truth. For each condition, a `$+$' marker is placed to represent the average performance with standard deviation. (e) Average precision, recall, and F1 scores achieved by the CNNs in the three conditions.}
    \label{fig:result}
\end{figure*}

\subsubsection{Efficacy of Eye-Gaze Labels for CNN Training} We evaluated three label generation methods for CNN training:

\begin{itemize}
    \item \textbf{Heuristic-Based}: Because the mitosis is usually stained brown in PHH3 immunohistochemistry, an HSV color detector was used to extract brown pigments in the images \cite{ruifrok2001quantification}. The locations of brown pigments were then used as the heuristic-based labels.
     \item \textbf{Eye-Gaze}: Generated from the eye-gaze data (Section \ref{s3p5}). We used the eye-gaze information from all participants (\ie $k$=14) to represent the best performance that the eye-gaze can provide.
    \item \textbf{Ground-Truth}: Previously annotated by pathologists.
\end{itemize}

For each labeling condition, we ran the CNN training experiments five times with the same procedures, training/validation split, and hyperparameters. We then compared the model predictions on the test WSI against the ground truth, and reported the precision, recall, and F1 scores.

%% file: 03_result.tex
\section{Results}

\subsection{Summary of Eye-Gaze Data}

In total, $800(\text{sequences/participant})\times 14 (\text{participants})=11,200$ eye-gaze sequences were collected. 470 sequences were discarded due to a high proportion of `N/A', low confidence, off-screen gaze location estimations, or data-logging failures. For the remaining 10,730 sequences, their average length was 157.77$\pm$0.77 (data points), \ie the average time each participant spent viewing each image was $\sim$2.63s. A more detailed distribution of participants' time consumption is shown in Figure \ref{fig:result}(a). We also collected participants' time consumption on annotating HPF images from the post-training survey, where they spent a longer average time of $\sim$10.27s on each image. Although a direct comparison of the two time consumption is not fair due to varying experimental setups, this observation nonetheless highlights the potential for reduced time cost in collecting eye-gaze data.

\subsection{Quality of Eye-Gaze Labels}
\label{s4p2}
Figure \ref{fig:result}(b--d) shows the precision, recall, and F1 scores of eye-gaze labels as a function of group size $k$. These subfigures reveal the following two trends:

\begin{enumerate}
    \item According to Figure \ref{fig:result}(b), there was a noticeable improvement in precision as $k$ increased (from 0.321 at $k$=3 to 0.679 at $k$=14). This improvement could be attributed to the reduction of random false-positive fixations as more participants were included.
    \item As shown in Figure \ref{fig:result}(c), the variation in the recall of eye-gaze labels decreased as $k$ increased, suggesting that eye-gaze labels from larger groups were more robust and generally less affected by the choice of participants.
\end{enumerate}

The highest quality eye-gaze labels were generated at $k$=14, with a precision of 0.679, a recall of 0.858, and an F1 score of 0.758. As a comparison, labels generated through the heuristic-based approach had a precision of 0.598, a recall of 0.966, and an F1 score of 0.739. Although the heuristic-based approach could identify almost all true-positive mitoses according to their color, it was prone to high false-positive rates due to the color similarity of non-mitoses artifacts. Therefore, it may cause the AI to learn from the spurious correlation and degrade its performance.

\subsection{Evaluation of CNNs Trained with Eye-Gaze Labels}

Figure \ref{fig:result}(e) shows the precision-recall ranges for the EfficientNet-b3 CNNs trained with heuristic-based labels, eye-gaze labels ($k$=14), and ground truth. The ground-truth-based CNNs had the highest performance, closely followed by eye-gaze-based CNNs: The precision-recall curve's upper bound for eye-gaze-based CNNs briefly overlapped with the lower bounds of ground-truth-based CNNs, indicating a close match in certain threshold ranges. In contrast, the performance of heuristic-based CNNs was significantly lower, with no overlap with the other two methods. These results suggest that eye-gaze data can retain more meaningful information for CNN training compared to heuristic-based methods, resulting in improved CNN performance.

As shown in Figure \ref{fig:result}(f), the precision achieved by eye-gaze-based CNNs was comparable to that of the ground-truth-based CNNs (0.83 \vs 0.85). Nonetheless, there was still a gap in recall when comparing eye-gaze-based CNNs against ground-truth-based CNNs (0.61 \vs 0.72).

%% file: 04_discussion.tex
\section{Conclusion}

This work introduces and evaluates a proof-of-concept pipeline that enables CNNs to learn from users' eye-gaze data, focusing on a pathology task of mitosis detection. To distill information from the noisy eye-gaze data, we extracted consistent eye-fixation areas observed by groups of users and employed them as pseudo labels. Compared to labels generated from the heuristic-based approach, eye-gaze data can provide more meaningful information to facilitate CNN model training. However, there was still a performance gap in recall between CNNs trained with eye-gaze labels and those with ground truth. Future research may close the gap by enhancing the quality of eye-gaze-based labels. For instance, one potential approach is to use heuristic-based labels refined by eye-gaze hotspots as a verification, which can achieve more accurate label localization and reduce false-positive errors. Another limitation of this study is the use of non-medical participants. In light of this, we encourage future research to expand the eye-tracking study to pathologists and further validate the results of this study.